\documentclass[11pt]{article}

\usepackage[letterpaper,margin=1in]{geometry}
\usepackage[T1]{fontenc}
\usepackage[utf8]{inputenc}
\usepackage{lmodern}
\usepackage{microtype}
\usepackage{doi}
\usepackage{booktabs}
\usepackage{placeins}
\usepackage{amsmath}
\usepackage{authblk}
\usepackage{graphicx}
\usepackage{subcaption}
\usepackage{tikz}
\usepackage{svg}
\usepackage{xurl}
\usepackage{hyphenat}
\usepackage[
  articletitle=true,
  doi=true
]{achemso}

\usepackage{hyperref}

\renewcommand{\doi}[1]{%
  \href{https://doi.org/#1}{\nolinkurl{#1}}%
}

\usetikzlibrary{
  arrows.meta,
  positioning,
  fit,
  shapes.geometric
}

\hypersetup{
  colorlinks=true,
  linkcolor=blue,
  citecolor=blue,
  urlcolor=blue
}

\title{Agent-MD: Selective LLM Intervention with Event-Driven Escalation for Stateful GCMC--MD Campaigns}
\author[1]{Yijie Wang}
\author[1]{Zhen-Yu Yin\thanks{Corresponding author: zhenyu.yin@polyu.edu.hk}}
\author[2]{Zhenheng Tang}
\author[3]{Xiaowen Chu}

\affil[1]{Department of Civil and Environmental Engineering, The Hong Kong Polytechnic University, Hong Kong SAR, China}
\affil[2]{Department of Computer Science and Engineering, The Hong Kong University of Science and Technology, Hong Kong SAR, China}
\affil[3]{Thrust of Artificial Intelligence, The Hong Kong University of Science and Technology (Guangzhou), Guangzhou, China}
\date{}

\begin{document}

\maketitle

\begin{abstract}
Long-running molecular simulation campaigns require repeated continuation from saved states, provenance-aware progression, adaptive assessment, and occasional interpretation of workflow conditions that cannot be resolved safely by fixed rules. Here, we present Agent-MD, a framework that places large language model (LLM) reasoning selectively at campaign construction and event-triggered review, while routine simulation, analysis, continuation, archiving, and state progression are handled by a persistent rule-based campaign agent using approved policies and explicit state records. Agent-MD was demonstrated in a grand canonical Monte Carlo–molecular dynamics (GCMC–MD) water-vapor desorption campaign comprising five montmorillonite systems and three sequential relative-humidity states (RH = 0.9 → 0.3 → 0.1). Across 15 system–RH states, the workflow completed 120 segmented simulation cycles with state-specific sampling lengths and provenance-aware restart inheritance. Routine production required no live reasoning-agent invocation, while one state reached a review boundary; two preserved incidents were subsequently evaluated through blinded reasoning-agent replay, which identified the underlying workflow problems and recommended appropriate follow-up actions. The simulations also revealed distinct composition-dependent low-RH responses, with Ca-bearing montmorillonite retaining more interlayer water and maintaining a larger basal spacing than the Na- and K-bearing systems, while the highest-charge Na system retained more residual water under dry conditions. These results demonstrate that long-running scientific workflows need not place every operation inside an LLM reasoning loop: selective reasoning can instead be combined with deterministic execution, structured evidence, and validated control handoffs to provide reproducible and auditable agent-assisted molecular simulation.
\end{abstract}
\section{Introduction}
Molecular dynamics (MD) simulations have become central tools for resolving molecular-scale mechanisms and predicting the structural, thermodynamic, and transport properties of chemical, material, and biological systems \cite{hollingsworth2018md,craven2025mosdef}. However, a reliable MD study involves considerably more than executing a trajectory. Researchers must construct and validate molecular models, select force fields and numerical settings, prepare engine-specific inputs, submit and monitor computational jobs, manage restart files, analyze evolving outputs, and preserve the provenance linking each result to its inputs and execution history \cite{adorf2018signac,bereau2024martignac,craven2025mosdef}. Coordinating these tasks across specialized simulation software, scripts, and file systems requires substantial technical expertise and repeated manual supervision. As simulations become longer and more interconnected, manual coordination also increases the risk of inconsistent parameters, incorrect restart selection, lost workflow context, and incomplete records.

These challenges make molecular simulation a natural setting for artificial intelligence (AI) agents. By interpreting high-level scientific objectives and invoking established domain tools, an agent can connect molecular-model preparation, input generation, simulation execution, monitoring, analysis, and reporting within a coordinated workflow. This mode of operation has the potential to lower the practical barrier to specialized simulation software, reduce repetitive supervision and record keeping, and make computational procedures more explicit and reproducible, provided that tool use and workflow decisions remain traceable. Recent studies have begun to demonstrate this potential. MDCrow employs a large language model (LLM)-guided reasoning--action loop to select expert-designed tools for molecular dynamics setup, execution, and analysis \cite{mdcrow}. Ding et al. combine global supervision, domain-specific skills, and high-performance computing execution in an OpenClaw-based computational chemistry architecture \cite{openclawchem}. Related agent-assisted workflows have automated metal--organic framework simulations and coordinated large-scale materials-screening tasks \cite{simmof,multiagentgcmc}. Together, these studies show that AI agents can act as interfaces and coordinators across specialized scientific software rather than serving only as assistants for isolated code or input-file generation.

Despite this progress, existing agent-assisted studies have mainly demonstrated tool selection, bounded workflow execution, or coordination across predefined simulation tasks \cite{mdcrow,openclawchem,simmof,multiagentgcmc}. Molecular simulation also encompasses long trajectories, adaptive sampling, and multistage campaigns in which subsequent calculations are selected or initialized using results from earlier stages \cite{shaw2010,kleiman2023adaptive,hruska2020extasy}. In these campaigns, the required sampling effort may not be known in advance, each simulation state may require repeated cycles of execution and analysis, and later states may depend on configurations accepted at earlier conditions. Scientific workflow systems have therefore emphasized explicit task dependencies, persistent metadata, and provenance-aware execution as foundations for reproducible computational research \cite{adorf2018signac,jain2015fireworks,pizzi2016aiida,huber2020aiida,bereau2024martignac}. For such workflows, completion of a submitted job is not necessarily equivalent to completion of the corresponding scientific state: multiple observables may need to be assessed, an authoritative restart or archive must be selected, and the relationship between predecessor and successor states must remain consistent. Many continuation, analysis, and archiving operations can be handled by explicit rules, whereas unexpected failures, ambiguous stabilization behavior, or provenance conflicts may require flexible interpretation. How and when an LLM-based agent should intervene in this mixture of routine and uncertain decisions, while established scientific tools retain responsibility for reliable computation and analysis, remains insufficiently explored.

Adsorption and desorption isotherm simulations provide a representative example of such long-running, state-dependent campaigns. A complete isotherm is assembled from simulations performed across a sequence of pressures or humidities rather than from a single calculation, and each condition requires sufficient sampling and an explicit assessment of whether the corresponding state is ready for reporting or progression \cite{dubbeldam2016,wang2023betgab}. In path-dependent adsorption or desorption studies, an accepted configuration at one condition may be used to initialize the next, making restart selection and predecessor-state provenance part of the scientific procedure. Structural relaxation, molecular uptake, and spatial redistribution may proceed differently across conditions, and different observables may not stabilize simultaneously; sorption hysteresis and structural transitions further increase this variability \cite{wang2025ca}. Constructing such an isotherm therefore involves repeated simulation, monitoring, continuation, state assessment, archiving, and progression, together with substantial manual judgment and record keeping. Agent-assisted execution offers considerable potential to reduce this repetitive supervision while preserving the validated simulation and analysis methods on which the scientific conclusions depend.

Montmorillonite provides a scientifically important and technically demanding test case. Hydration of its interlayers drives changes in water organization and clay-sheet spacing, producing swelling and shrinkage that influence expansive-soil deformation, borehole stability, and the performance of clay barriers in geoenvironmental and nuclear-waste applications \cite{amarasinghe2012clayfluid,teichmcgoldrick2015swelling,yotsuji2021cation}. These responses depend strongly on exchangeable-cation identity and layer charge; the latter is closely related to cation exchange capacity (CEC) and the population of charge-balancing interlayer ions \cite{zhang2014hydration,yang2019layercharge,tambach2004swelling}. In experiments, isolating the effect of layer charge is difficult because charge, CEC, exchangeable-ion population, and mineral structure are often coupled, and controlled manipulation requires dedicated chemical treatments \cite{ma2025layercharge}. Molecular dynamics and GCMC--MD provide complementary mechanistic tools because molecular composition and layer charge can be prescribed while interlayer water, ion organization, sorption, and clay-sheet separation are resolved directly \cite{teichmcgoldrick2015swelling,wang2025ca}. Montmorillonite water-vapor desorption therefore offers both a demanding stateful workflow and a meaningful physical problem for evaluating agent-assisted molecular simulation.

We therefore developed Agent-MD as a framework for selective LLM intervention with event-driven escalation in long-running molecular simulation campaigns. The framework assigns campaign construction and unresolved review to an LLM-based reasoning agent, while a persistent rule-based campaign agent manages routine production through approved policies, validated scientific computation and analysis tools, and explicit state and provenance records. Recommendations from the reasoning agent are validated before they can affect execution, allowing flexible interpretation to re-enter only when routine rules cannot safely determine the next action. We demonstrate this architecture in a GCMC--MD water-vapor desorption campaign comprising five montmorillonite systems and three sequential RH states, with adaptive continuation, accepted-state inheritance, and event-triggered review. The workflow is evaluated through its operational record, an authentic production incident, a provenance conflict, and blinded replay of preserved cases. In parallel, the simulations examine how exchangeable-cation identity and layer charge influence residual hydration, water partitioning, and clay-sheet contraction during desorption. This dual demonstration establishes Agent-MD both as a practical workflow architecture for reliable and auditable long-running simulation and as a tool for addressing a substantive molecular-scale materials problem.
\section{Agent-MD Framework}

\subsection{Overall Architecture and Agent Roles}

Agent-MD consists of a human researcher, an LLM-based reasoning agent, a persistent rule-based campaign agent, scientific computation and analysis tools, and persistent campaign state and provenance records. The human researcher defines the scientific objective and retains responsibility for approving the campaign and any changes that require scientific judgment. The reasoning agent is used for tasks requiring flexible interpretation, principally campaign construction and review of unresolved production states. These two functions are performed by the same LLM-based reasoning role at different intervention points. The rule-based campaign agent manages routine production by reading the approved campaign specification and current campaign state, selecting the next permitted action, invoking the appropriate scientific tools, and updating the campaign records. The distinction between the two agent roles therefore lies primarily in how their actions are selected: the reasoning agent interprets bounded scientific or workflow problems, whereas the rule-based campaign agent follows approved policies and recorded state.

\begin{figure}[t]
    \centering
    \includegraphics[width=0.98\linewidth]{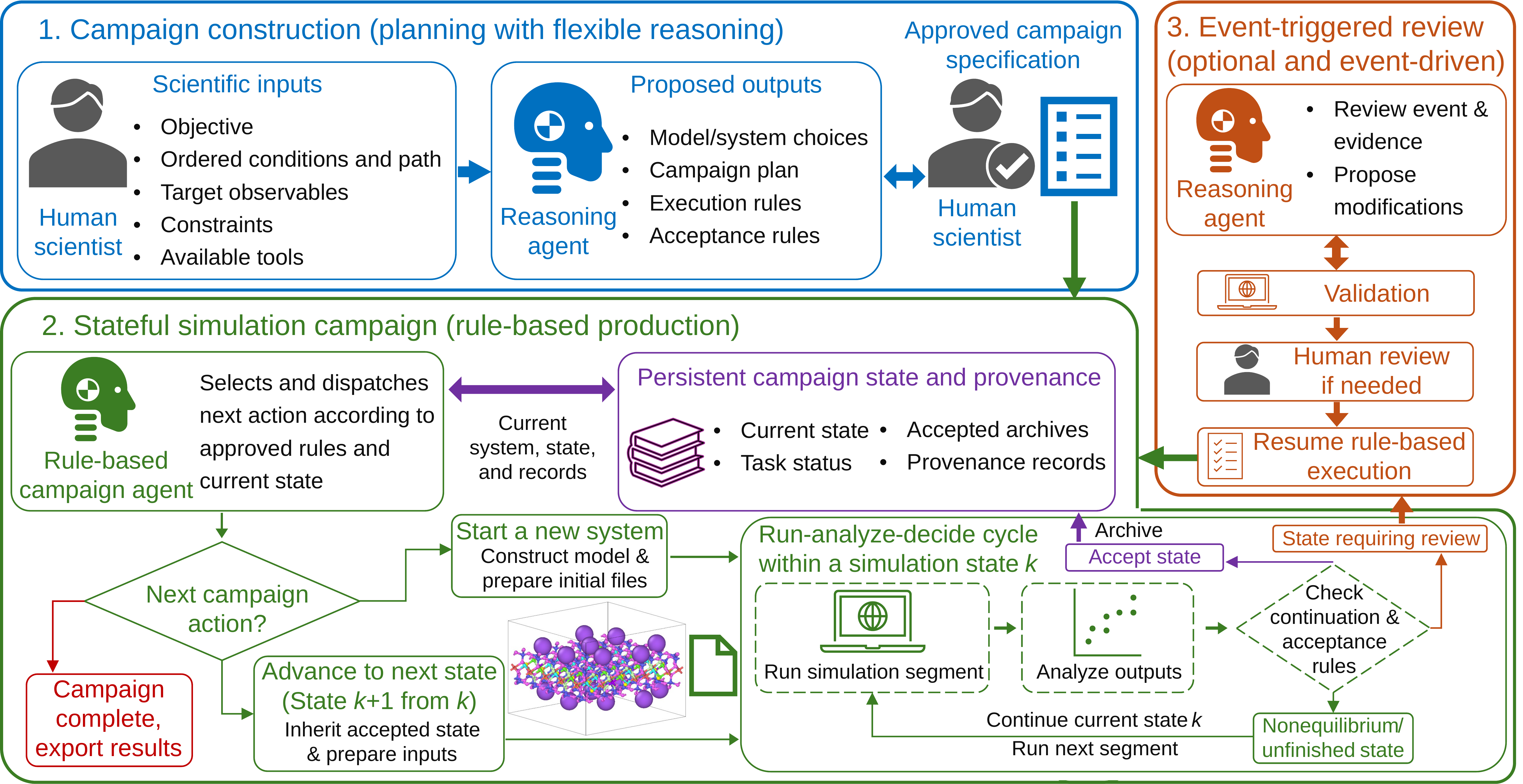}
    \caption{General architecture of Agent-MD for stateful simulation campaigns. Campaign construction produces a human-approved specification, the rule-based campaign agent manages routine stateful production, and unresolved states can be routed to event-triggered review before validated actions return to production.}
    \label{fig:agent_md_architecture}
\end{figure}

As shown in Figure~\ref{fig:agent_md_architecture}, the architecture is organized into three connected stages. Campaign construction begins with researcher-defined inputs, including the scientific objective, prescribed conditions or path, target observables, constraints, and available tools. The reasoning agent translates these inputs into proposed molecular-system choices, a campaign plan, and execution and acceptance rules. The researcher then reviews and approves the proposal. The resulting campaign specification forms an explicit boundary between flexible planning and production execution: it can be inspected and validated before being used by the rule-based campaign agent. During routine production, the reasoning agent is not continuously active; it re-enters the workflow only when an unresolved state is routed to event-triggered review.

Production is organized as a stateful simulation campaign. A molecular system denotes a distinct model or composition, whereas a simulation state denotes that system at one prescribed condition together with its current configuration, sampling progress, analysis and acceptance status, and provenance. Starting a new molecular system requires model construction and preparation of its initial simulation files. By contrast, progression from state \(k\) to state \(k+1\) uses the accepted predecessor state to prepare the inputs for the next prescribed condition. New-system construction and accepted-state inheritance are therefore treated as distinct workflow actions.

Within each simulation state, the rule-based campaign agent coordinates a segmented run--analyze--decide cycle. After each simulation segment, automated routines analyze the outputs and apply the approved continuation and acceptance rules. A state that requires additional sampling is continued from its current valid restart; an accepted state is archived together with its supporting records; and a state that cannot be safely resolved by the approved rules is routed to event-triggered review. Completion of an individual simulation segment therefore does not by itself imply completion of the corresponding simulation state.

During event-triggered review, the reasoning agent receives a structured review event and an associated evidence bundle and proposes a response or modification. The proposal is validated before it can affect production, and human review is used when scientific approval or further interpretation is required. A validated action is returned to the rule-based campaign agent, which resumes execution and records the outcome. Persistent campaign state and provenance connect campaign construction, production, and review, preserving the information required to continue, reproduce, and audit the campaign.

\subsection{Campaign Construction and Montmorillonite System Preparation}

To demonstrate how the general Agent-MD architecture can be instantiated in a scientific workflow, we applied it to a montmorillonite water-vapor desorption campaign. The human scientist prescribed the scientific objective, the ordered desorption path
\(\mathrm{RH}=0.9\rightarrow0.3\rightarrow0.1\), the comparison of exchangeable-cation and layer-charge effects, and the target observables. This RH range is sufficient to observe the dehydration and shrinkage of montmorillonite \cite{ferrage2007investigation,wang2025ca}. The observables included total water content, the partitioning of water between the interlayer and external-surface regions, and the evolution of clay-sheet spacing (i.e., basal spacing). The scientific instruction therefore defined the comparison to be performed and the conditions to be followed, but did not enumerate the exact cation species, layer-charge levels, or molecular systems.

During campaign construction, the reasoning agent used the scientific objective together with the available model-construction, simulation, and analysis tools to propose a molecular-system matrix, an ordered state sequence, execution rules, and acceptance rules. The proposed choices were restricted to systems that could be prepared and simulated using the available ClayCode- and ClayFF-based workflow \cite{pollak2024claycode,cygan2004clayff}. Low-level molecular and numerical parameters that were already fixed by the validated preparation and simulation protocols were not independently selected by the reasoning agent. The human scientist reviewed and approved the proposal before it was converted into the executable campaign specification. The reasoning agent therefore assisted with scientific organization and system selection, but neither approved the campaign independently nor directly executed the simulations.

\begin{figure}[htb]
    \centering
    \includegraphics[width=0.98\linewidth]{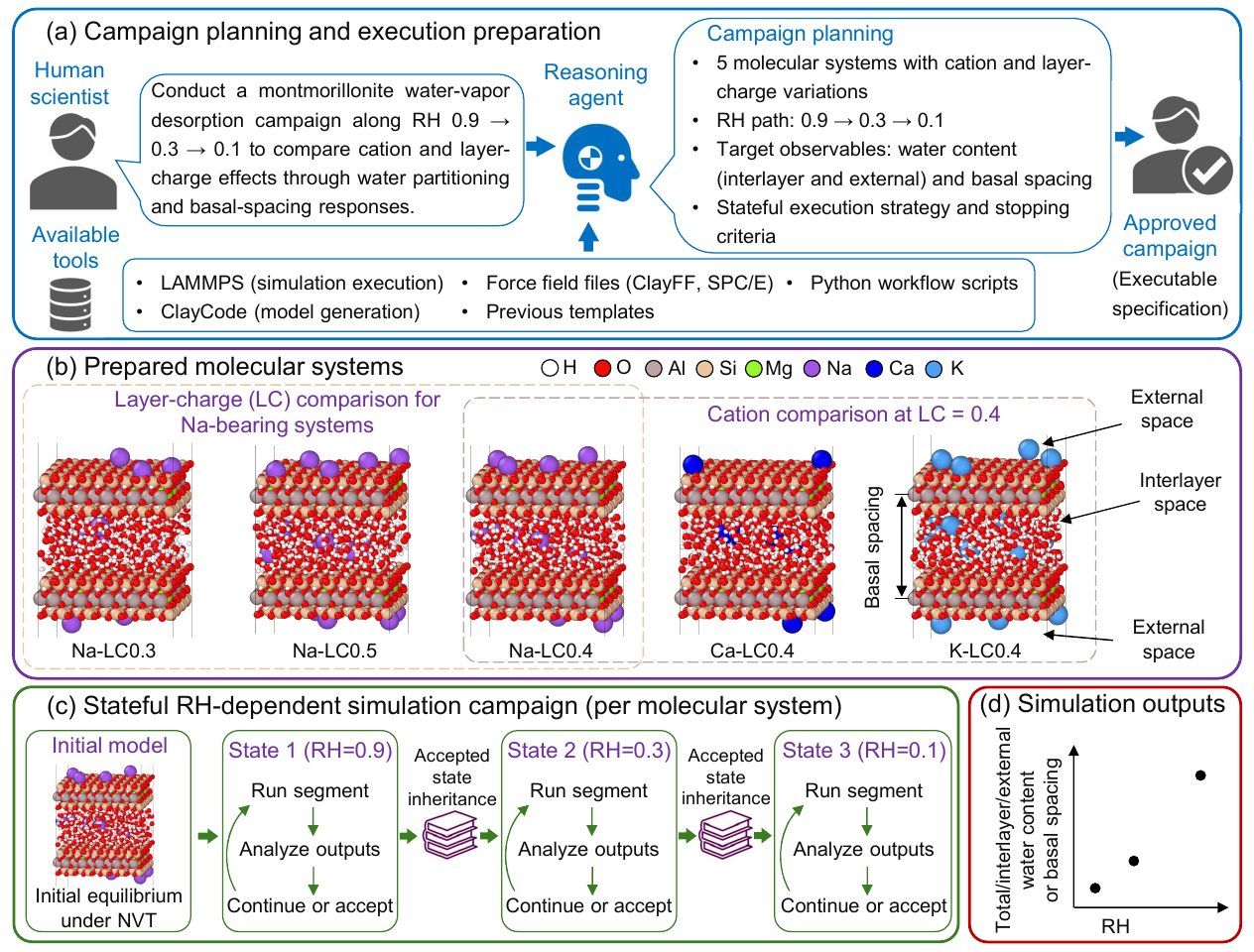}
    \caption{Instantiation of Agent-MD for the montmorillonite water-vapor desorption campaign. A researcher-defined objective and the available scientific tools were translated into a human-approved campaign specification. Five prepared molecular systems were then advanced through the prescribed humidity sequence using state-specific run--analyze--decide cycles and accepted-state inheritance.}
    \label{fig:campaign_instantiation}
\end{figure}

Figure~\ref{fig:campaign_instantiation} shows how this planning stage was connected to molecular-system preparation and stateful production. The approved design contained two complementary comparisons. To examine exchangeable-cation effects at a common layer-charge level, Na$^{+}$, K$^{+}$, and Ca$^{2+}$ bearing montmorillonite systems were prepared at layer charge (LC) \(=0.4\). To examine LC-dependent behavior within the Na-bearing series, systems with LC \(=0.3\), \(0.4\), and \(0.5\) were prepared. The Na-LC0.4 system was shared between the two comparisons, giving five distinct molecular systems rather than a full factorial combination. Here, LC denotes the magnitude of the nominal negative layer charge, expressed in units of the elementary charge \(e\) per \(\mathrm{O}_{20}(\mathrm{OH})_{4}\) structural unit. Combining the five systems with the three prescribed RH conditions produced a planned campaign of 15 system--RH simulation states. Because the number of charge-balancing Na ions changes with layer charge, the Na series represents the combined variation of layer charge and its corresponding compensating-ion population rather than a fixed-ion-population comparison.

The approved campaign design was materialized as a campaign specification, five system-specific case records, a machine-readable task plan, and a persistent campaign-state record. The campaign specification encoded the molecular-system matrix, prescribed RH sequence, target observables, continuation and acceptance policies, and state-transition rules. The case records stored system-specific composition, layer charge, exchangeable-cation identity, model size, and preparation parameters. The task plan represented the dependencies among molecular-system construction, initial preparation, RH-specific production, analysis, acceptance, archiving, and progression to the next condition. The persistent campaign-state record tracked the status of these tasks and the current position of each system along the prescribed RH path. Together, these records converted the approved scientific plan into inspectable and executable inputs for the rule-based campaign agent.

Following approval, molecular-system preparation was carried out through established scientific tools. The rule-based campaign agent translated the approved system definitions, including layer charge and exchangeable-cation identity, into the mineral-composition and charge-balancing-ion information required by ClayCode. System-specific ClayCode input files were then generated programmatically from the campaign records and validated templates. ClayCode constructed the atomistic clay structures from these mineral-composition specifications and substitution patterns, while ClayFF and SPC/E supplied the force-field descriptions for the clay, exchangeable ions, and water \cite{pollak2024claycode,cygan2004clayff,berendsen1987spce}. The rule-based campaign agent then invoked the preparation workflow and converted the resulting structure and topology files into LAMMPS-ready molecular data files \cite{plimpton1995lammps,thompson2022lammps}. LAMMPS inputs were generated from templates adapted from previously validated clay--water GCMC--MD protocols \cite{wang2025ca}. System- and state-specific values, including molecular-system identity, RH-dependent conditions, restart paths, output names, monitored quantities, and production-segment settings, were inserted programmatically rather than edited separately for each case.

Automated checks were applied before a prepared system could enter production. These checks included charge balance, integer exchangeable-cation counts, expected ion partitioning, exchangeable-ion group definitions, molecule identifiers, atom types, atomic charges, required input files, and compatibility with the water template. Preparation summaries and validation reports were recorded together with the molecular inputs and initial state information. The resulting artifacts provided the rule-based campaign agent with an auditable starting point for production and preserved the connection between the approved campaign design and each prepared molecular system.

\subsection{Stateful GCMC--MD Production and Analysis}

\subsubsection{Simulation Setup and State Initialization}

Following campaign approval and molecular-system preparation, the rule-based campaign agent managed production over the five-system, three-RH state matrix. Each montmorillonite model consisted of a \(5 \times 4\) lateral replication containing two clay sheets, one interlayer region, and two external-surface regions within a common simulation cell of approximately \(25.8 \times 35.864 \times 86.7~\text{\AA}^{3}\). Note that sufficient vacuum layers are provided in the height direction to facilitate water adsorption on the external surfaces. The Na--LC0.3, Na--LC0.4, Na--LC0.5, K--LC0.4, and Ca--LC0.4 systems contained 12, 16, 20, 16, and 8 exchangeable cations, respectively. These ion populations balanced the specified structural charge, and automated preparation checks confirmed residual simulation-cell charges below \(0.004e\).

ClayFF was used to describe the clay framework and exchangeable ions, while water was represented by the SPC/E model \cite{cygan2004clayff,berendsen1987spce}. GCMC--MD production was performed with LAMMPS using real units and periodic boundary conditions in all three directions \cite{plimpton1995lammps,thompson2022lammps}. The MD timestep was \(1~\mathrm{fs}\), and both the Lennard--Jones cutoff and real-space Coulomb cutoff were \(12~\text{\AA}\). Long-range electrostatic interactions were evaluated using the particle--particle particle--mesh method with a relative accuracy of \(10^{-4}\). Water molecules and exchangeable ions were maintained at \(300~\mathrm{K}\) using a thermostat damping constant of \(100~\mathrm{fs}\), and the geometry of each SPC/E water molecule was constrained using SHAKE. Each clay sheet was treated as a rigid body that could translate along the \(z\) direction, allowing the interlayer separation to respond to changes in hydration while preserving the clay-sheet structure. These settings followed the previously validated clay--water GCMC--MD protocol used as the basis for the campaign \cite{wang2025ca}.

Before GCMC--MD production, each newly constructed molecular system was equilibrated under NVT conditions at \(300~\mathrm{K}\), without water insertion or deletion moves. The resulting restart was recorded as the authoritative initialization artifact for the first prescribed state at RH \(=0.9\). Subsequent RH states were initialized from the authoritative archived restart of the accepted predecessor state: RH \(=0.9\) initialized RH \(=0.3\), and RH \(=0.3\) initialized RH \(=0.1\). Pre-GCMC equilibration and accepted-state inheritance therefore provided two distinct initialization routes, and successor states were not initialized from arbitrary recent files in the working directory.

For each RH condition, the imposed water-vapor pressure was calculated as \(p=\mathrm{RH}\,p_{\mathrm{sat}}\), where \(p_{\mathrm{sat}}\) is the saturation vapor pressure of the SPC/E water model at \(300~\mathrm{K}\), approximately \(1.0~\mathrm{kPa}\) \cite{brochard2021swelling}. The production protocol requested 100 molecular insertion or deletion attempts every 1000 MD steps. Inserted water molecules used the same SPC/E topology and SHAKE constraints as the water molecules already present in the system.

\subsubsection{Segmented Execution, Analysis, and State Progression}

Once initialized, each simulation state was advanced through a sequence of finite GCMC--MD segments. Each completed segment generated monitor data, restart files, LAMMPS logs, and structured run-status records, after which automated analysis was performed. The total sampling length was not fixed uniformly across the campaign because the sampling required for stabilization could differ among molecular systems and RH conditions. Instead, the state was reassessed after each segment through the run--analyze--decide cycle.

The monitored observables were total water, interlayer and external-surface water populations, and the basal-spacing proxy, as illustrated in Figure~\ref{fig:campaign_instantiation}(b). Water populations are reported as numbers of molecules per simulation cell. The basal-spacing proxy was calculated as the separation between the dynamically updated mean \(z\) positions of the two rigid clay sheets.

After each production segment, the analyzer calculated coefficients of variation and linear trends for total, interlayer, and external-surface water, together with the linear trend in the basal-spacing proxy, using the most recent \(1.0\times10^{6}\) MD steps. The corresponding coefficients-of-variation thresholds for total, interlayer, and external-surface water were 0.03, 0.05, and 0.08, respectively. Additional checks covered water-partition consistency, exchangeable-ion counts, temperature stability, simulation health, and restart provenance.

The resulting metrics were evaluated against the approved acceptance criteria encoded in the machine-readable campaign specification. These criteria were used as operational indicators for state assessment rather than as proof of strict thermodynamic equilibrium.

The rule-based campaign agent used the analysis results and recorded workflow state to select the next action. When the acceptance criteria were not yet satisfied but the simulation remained healthy and further sampling was permitted, production continued from the current valid restart. When the criteria and supporting checks were satisfied, the state was accepted and the authoritative restart was archived together with the corresponding analysis and provenance records. The accepted archive then became available for initialization of the next prescribed RH state. When the approved rules could not safely determine whether to continue or accept the state, production paused and the unresolved state was transferred to the event-triggered review procedure described in Section~2.4.

Because LAMMPS restarts preserve the cumulative simulation timestep, successor RH states generally did not begin at an absolute timestep of zero. Agent-MD therefore recorded both the absolute timestep and the RH-local steps accumulated after entry into the current RH condition. Analysis windows, continuation budgets, and state-specific sampling limits were evaluated using RH-local steps, whereas the absolute timestep was retained for restart continuity and provenance.

Scientific summaries were calculated over the final \(1.0\times10^{6}\) steps of each accepted state. Error bars represent temporal sample standard deviations within this window and therefore describe fluctuations along the sampled trajectory rather than uncertainty across independent simulation replicas.

\subsection{Event-Triggered Review and Decision Validation}

The reasoning agent was not invoked during routine simulation execution, continuation, acceptance, archiving, or RH progression. Event-triggered review was initiated only when the rule-based campaign agent reached a workflow state that could not be safely resolved using the approved rules. Events were generated at deterministic action boundaries, after a workflow action completed or failed, rather than through continuous LLM monitoring or periodic polling.

The current implementation defines five review-event identifiers: \texttt{CONVERGENCE\_ANOMALY}, \path{PROVENANCE_CONFLICT}, \texttt{UNKNOWN\_FAILURE}, \texttt{NEEDS\_REASONING}, and \texttt{BATCH\_COMPLETE}. These represent anomalous stabilization behavior, inconsistent state or restart provenance, an unclassified execution failure, a state requiring flexible interpretation, and a completed batch requiring additional review, respectively. Here, \path{BATCH_COMPLETE} denotes completion of a configured production batch for which an explicit post-batch review was requested, rather than routine completion of an accepted simulation state.

For each review event, the rule-based campaign agent constructed a focused evidence bundle from the persistent campaign state, provenance records, and relevant workflow artifacts. The evidence could include the action result and stop reason, molecular-system identity, RH condition, analysis summary, previous campaign decision, archive metadata, restart lineage, error signatures, and selected status or monitor records. The event description and supporting evidence were stored in \texttt{agent\_event.json} and \texttt{evidence\_bundle.json}. The same files could be inspected by either the reasoning agent or a human reviewer, providing a common and auditable review interface.

The optional reasoning agent was implemented through the Codex CLI and powered by GPT-5.5. When invocation was enabled and permitted by the escalation policy, the reasoning agent received the structured event and evidence bundle and returned a schema-constrained recommendation in \texttt{agent\_decision.json}. The reasoning agent could interpret the evidence and recommend a next action, but it did not directly modify simulation files, launch jobs, or alter scientific parameters.

Each recommendation was checked before it could affect production. Validation required consistency with the current event identifier, an allowed decision and recommended action, a valid confidence value, and compliance with the permitted scope of parameter changes. Parameter modification was restricted to no change or an adjustment of the subsequent segment length; decisions involving greater scientific or operational risk required human review. The validation outcome was recorded in \texttt{decision\_validation.json}.

Additional safeguards limited unnecessary or repeated reasoning-agent calls. Identical error fingerprints were deduplicated, calls were capped for repeated system--event combinations and for the campaign as a whole, and a recursion guard prevented an escalation action from triggering itself. If reasoning-agent invocation was disabled, unavailable, timed out, returned an invalid response, or failed to produce a decision, the workflow conservatively recorded a requirement for human review.

A validated recommendation was returned to the rule-based campaign agent, which executed the permitted response, such as resuming deterministic production, retrying a safe action, stopping the campaign, or waiting for human review. The resulting action and updated workflow state were then written back to the persistent campaign records, completing the event-triggered review loop.
\section{Results}

\subsection{Campaign Completion and Sampling Effort Across Simulation States}

All five molecular systems completed the prescribed water-vapor desorption path,
\(\mathrm{RH}=0.9\rightarrow0.3\rightarrow0.1\). The resulting 15 system--RH states were accepted and archived after satisfying the configured acceptance and provenance checks. Rather than assigning the same production length to every state, the rule-based campaign agent reassessed each state after every simulation segment and continued sampling when the approved criteria had not yet been satisfied.

Table~\ref{tab:production_summary} reports the number of completed simulation cycles, RH-local sampling effort, and final absolute timestep for each state. Each simulation cycle comprised one segmented GCMC--MD run followed by automated analysis and a decision to continue, archive, or request review. RH-local steps measure the sampling accumulated after entry into the current RH condition, whereas the final absolute timestep retains the counter inherited through the restart chain.

\begin{table}[tb]
  \centering
  \caption{State-specific production effort.}
  \label{tab:production_summary}
  \small
  \setlength{\tabcolsep}{5pt}
  \begin{tabular}{lrrrr}
    \toprule
    System &
    RH &
    \shortstack{Simulation\\cycles} &
    \shortstack{RH-local steps\\(\(\times10^{6}\))} &
    \shortstack{Final absolute timestep\\(\(\times10^{6}\))} \\
    \midrule
    Na--LC0.3 & 0.9 & 7  & 9.0  & 9.1  \\
    Na--LC0.3 & 0.3 & 9  & 18.0 & 27.1 \\
    Na--LC0.3 & 0.1 & 3  & 6.0  & 33.1 \\
    Na--LC0.4 & 0.9 & 6  & 8.0  & 8.1  \\
    Na--LC0.4 & 0.3 & 14 & 30.0 & 38.1 \\
    Na--LC0.4 & 0.1 & 1  & 2.0  & 40.1 \\
    Na--LC0.5 & 0.9 & 8  & 12.0 & 12.1 \\
    Na--LC0.5 & 0.3 & 10 & 20.0 & 32.1 \\
    Na--LC0.5 & 0.1 & 6  & 12.0 & 44.1 \\
    K--LC0.4  & 0.9 & 24 & 42.0 & 42.1 \\
    K--LC0.4  & 0.3 & 12 & 24.0 & 66.1 \\
    K--LC0.4  & 0.1 & 1  & 2.0  & 68.1 \\
    Ca--LC0.4 & 0.9 & 7  & 10.0 & 10.2 \\
    Ca--LC0.4 & 0.3 & 8  & 16.0 & 26.2 \\
    Ca--LC0.4 & 0.1 & 4  & 8.0  & 34.2 \\
    \bottomrule
  \end{tabular}
\end{table}

The required sampling differed substantially among the 15 states. RH-local sampling ranged from \(2.0\times10^{6}\) to \(42.0\times10^{6}\) simulation steps, with a median of \(12.0\times10^{6}\) steps. K--LC0.4 at \(\mathrm{RH}=0.9\) required the greatest sampling effort, reaching \(42.0\times10^{6}\) RH-local steps over 24 simulation cycles. Na--LC0.4 at \(\mathrm{RH}=0.3\) also required extended sampling, whereas several low-RH states were accepted after only \(2.0\times10^{6}\) additional steps. These differences show why a single fixed production length would not have been suitable for all systems and RH conditions.

\begin{figure}[!htbp]
  \centering
  \begin{subfigure}[b]{0.33\linewidth}
    \centering
    \includegraphics[width=\linewidth]{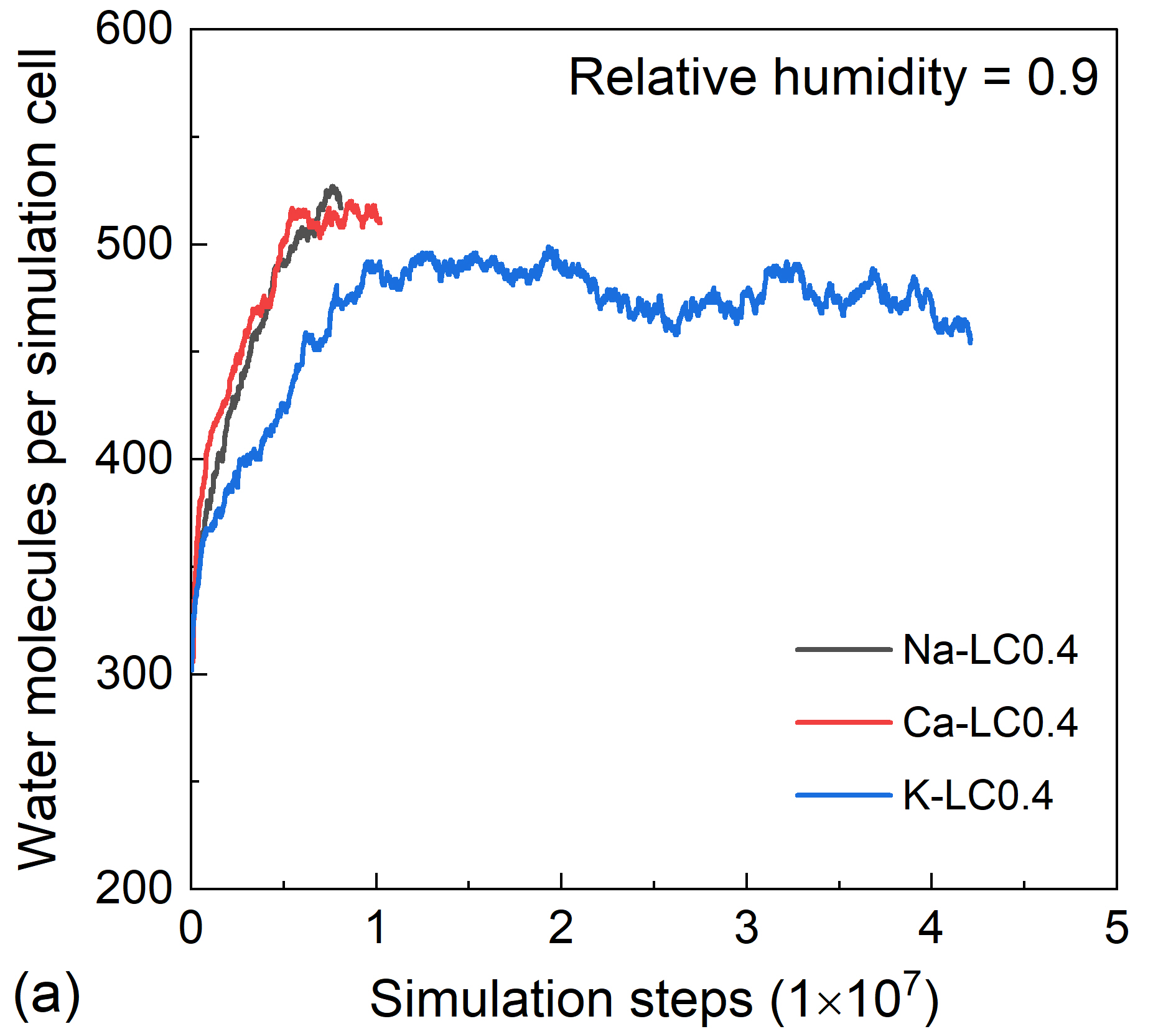}
    \label{fig:cation_hydration_trajectories_a}
  \end{subfigure}\hfill
  \begin{subfigure}[b]{0.33\linewidth}
    \centering
    \includegraphics[width=\linewidth]{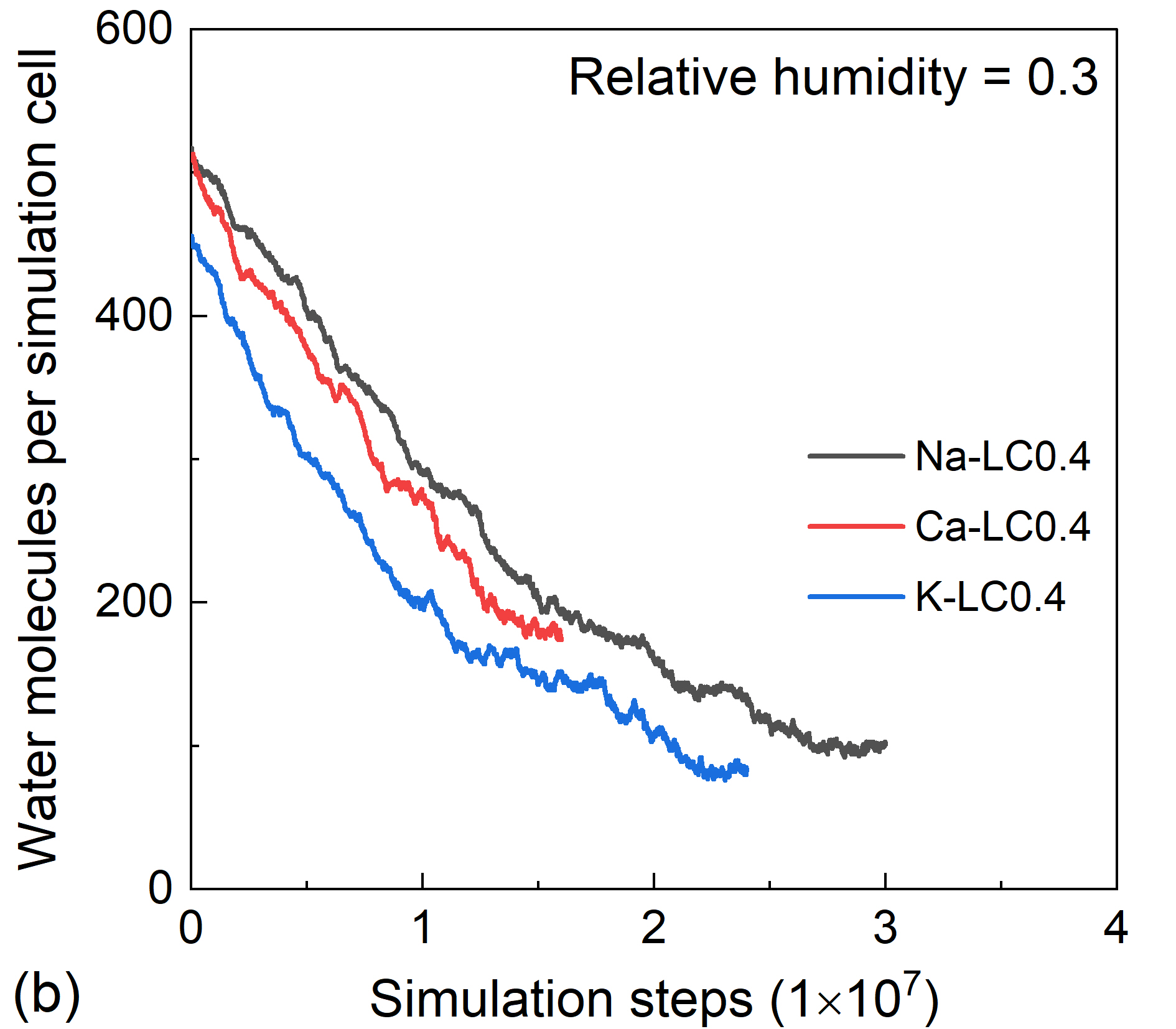}
    \label{fig:cation_hydration_trajectories_b}
  \end{subfigure}\hfill
  \begin{subfigure}[b]{0.33\linewidth}
    \centering
    \includegraphics[width=\linewidth]{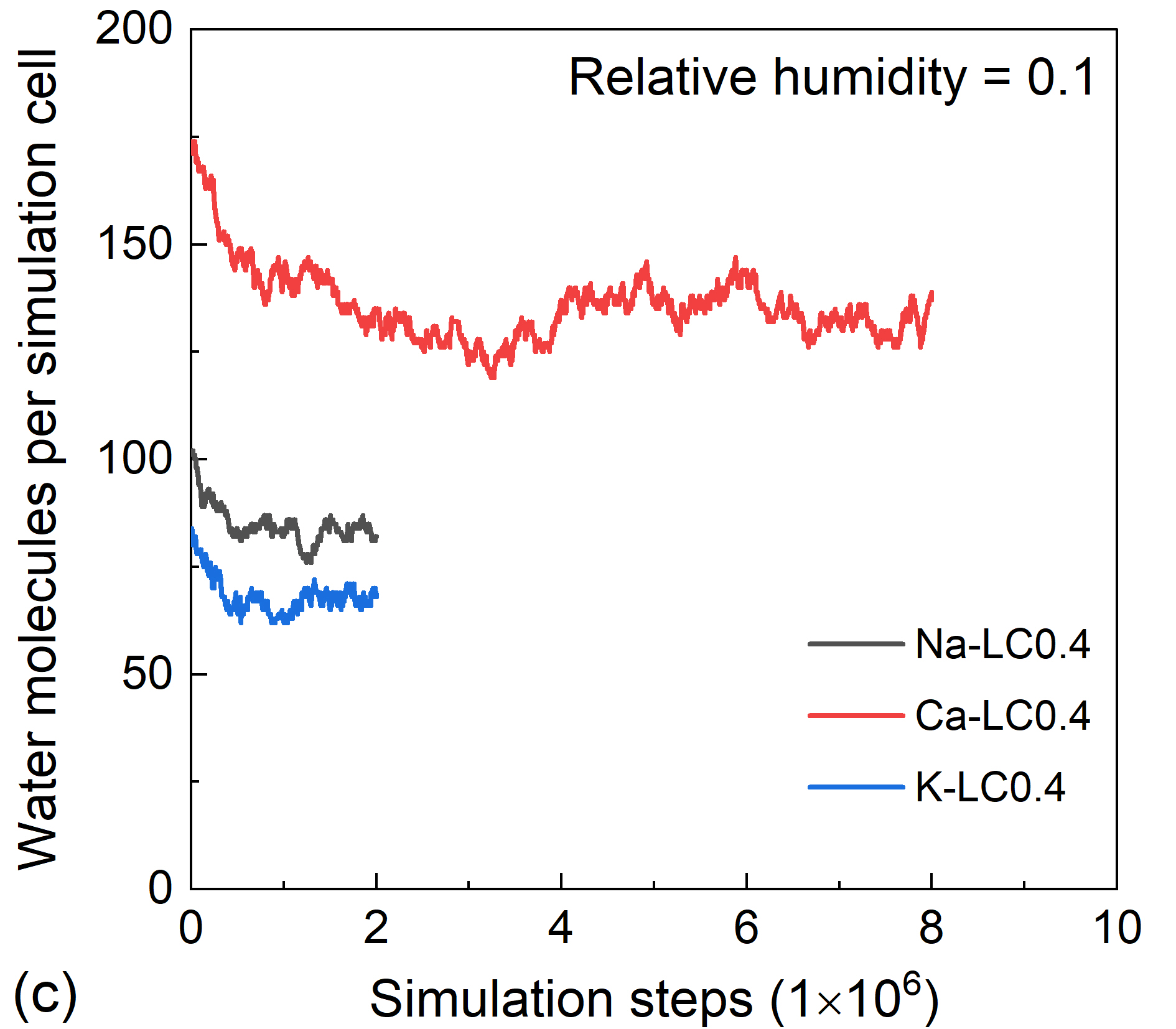}
    \label{fig:cation_hydration_trajectories_c}
  \end{subfigure}

  \caption{RH-local total-water trajectories for Na--LC0.4, K--LC0.4, and Ca--LC0.4 at (a) \(\mathrm{RH}=0.9\), (b) \(\mathrm{RH}=0.3\), and (c) \(\mathrm{RH}=0.1\).}
  \label{fig:cation_hydration_trajectories}
\end{figure}

\begin{figure}[!htbp]
  \centering
  \begin{subfigure}[b]{0.33\linewidth}
    \centering
    \includegraphics[width=\linewidth]{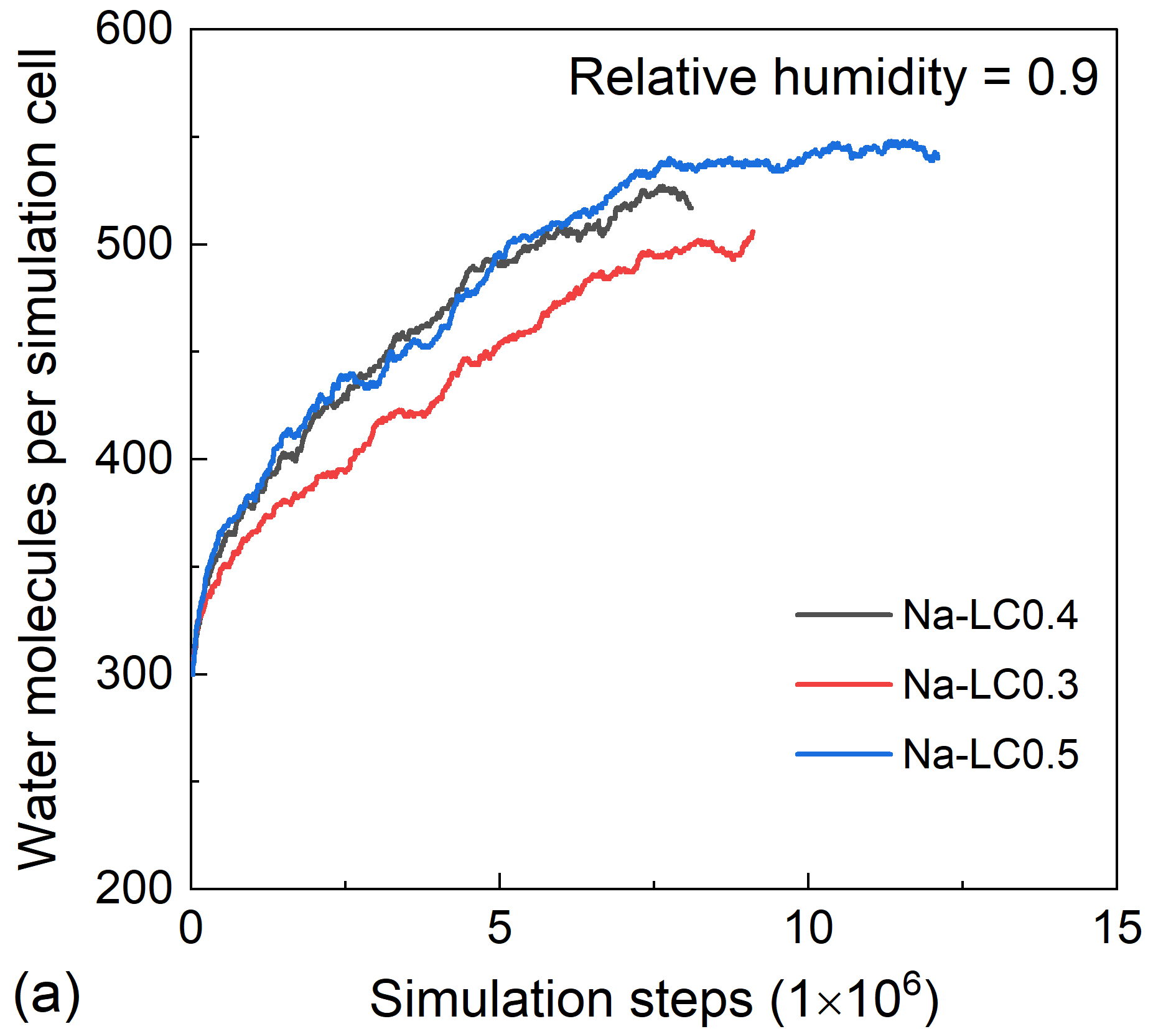}
    \label{fig:layer_charge_hydration_trajectories_a}
  \end{subfigure}\hfill
  \begin{subfigure}[b]{0.33\linewidth}
    \centering
    \includegraphics[width=\linewidth]{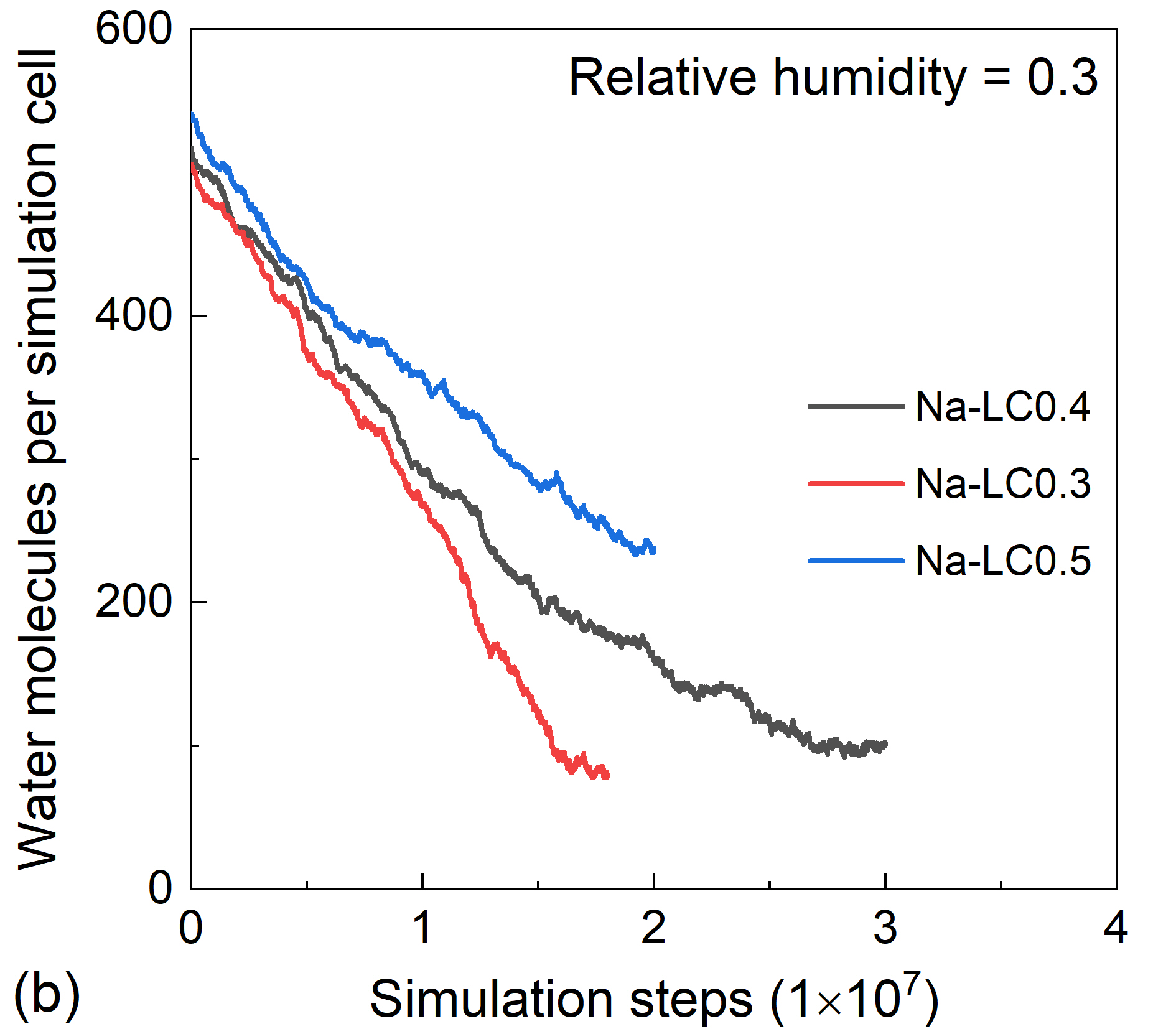}
    \label{fig:layer_charge_hydration_trajectories_b}
  \end{subfigure}
  \begin{subfigure}[b]{0.33\linewidth}
    \centering
    \includegraphics[width=\linewidth]{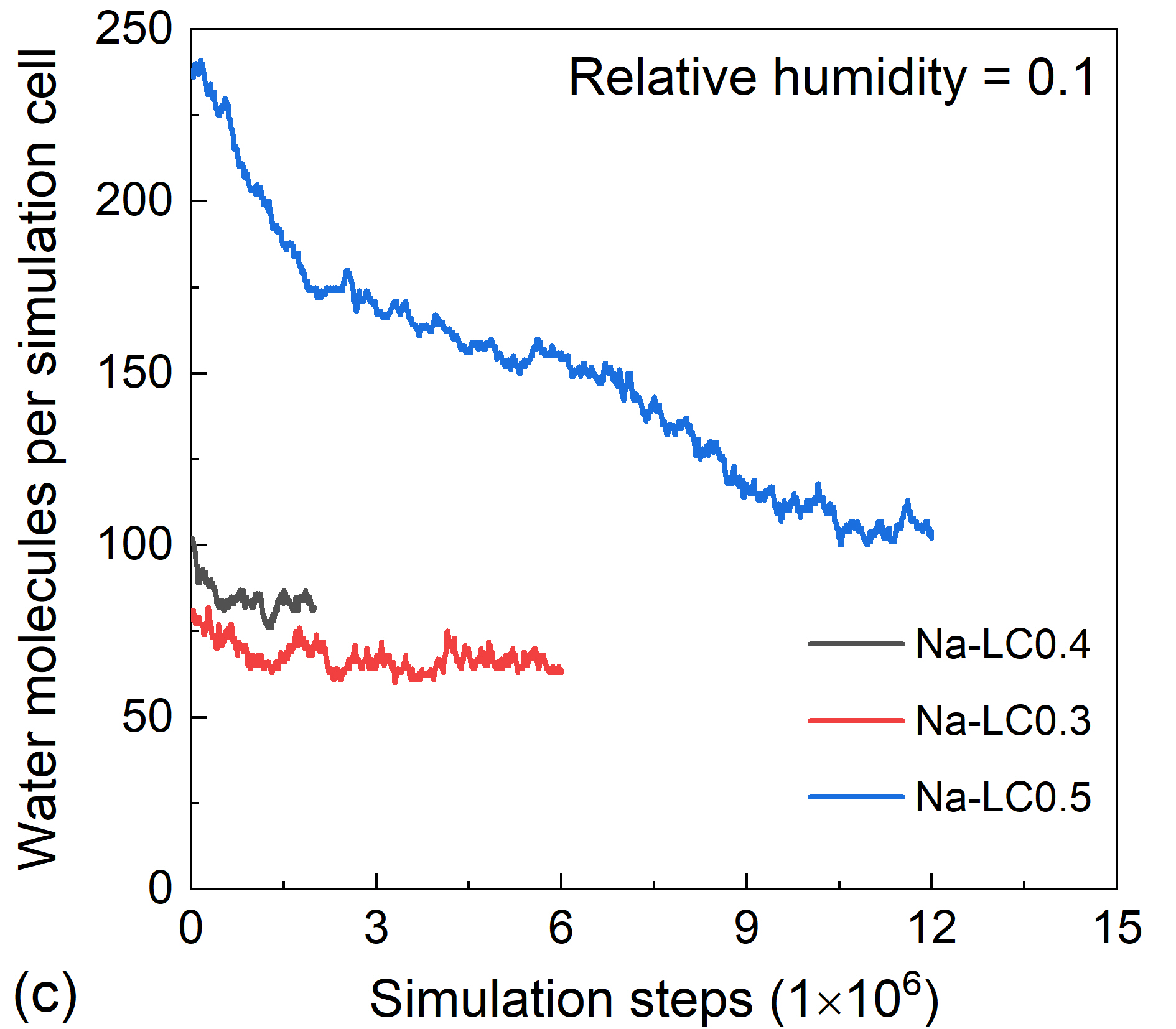}
    \label{fig:layer_charge_hydration_trajectories_c}
  \end{subfigure}
  \caption{RH-local total-water trajectories for Na--LC0.3, Na--LC0.4, and Na--LC0.5 at (a) \(\mathrm{RH}=0.9\), (b) \(\mathrm{RH}=0.3\), and (c) \(\mathrm{RH}=0.1\).}
  \label{fig:layer_charge_hydration_trajectories}
\end{figure}

Figures~\ref{fig:cation_hydration_trajectories} and~\ref{fig:layer_charge_hydration_trajectories} show how the water populations evolved during this state-specific sampling. The horizontal axis is the number of RH-local simulation steps. Differences in trajectory length therefore indicate how much sampling was performed before each state was accepted; they should not be interpreted as physical desorption times or intrinsic kinetic rates.

For the three LC0.4 systems, Figure~\ref{fig:cation_hydration_trajectories} shows that the required progression differed both among cations and among RH conditions. The long K--LC0.4 trajectory at \(\mathrm{RH}=0.9\) records repeated continuation before acceptance, whereas its \(\mathrm{RH}=0.1\) state required only one additional simulation cycle. The Na- and Ca-bearing systems followed different sequences of water loss and also reached acceptance after different RH-local sampling lengths.

The Na series displayed similarly varied sampling requirements (Figure~\ref{fig:layer_charge_hydration_trajectories}). Na--LC0.4 required \(30.0\times10^{6}\) RH-local steps at \(\mathrm{RH}=0.3\), whereas its \(\mathrm{RH}=0.1\) successor was accepted after \(2.0\times10^{6}\) steps. Na--LC0.5 required longer sampling than the other two Na systems at \(\mathrm{RH}=0.1\). The trajectories also show composition-dependent differences in retained water, which are compared using the accepted-state averages in Section~3.4.

\FloatBarrier

\subsection{Production Workload and Review Frequency}

Table~\ref{tab:operation_census} summarizes the principal workflow activity during formal production. The completed campaign contained 120 segmented GCMC–MD simulation cycles and 120 corresponding analysis–decision cycles. Every completed simulation segment was followed by automated analysis and a decision to continue the current state, accept and archive it, or request review. The workflow ultimately produced 15 accepted system–RH states and completed the 10 RH transitions required to advance five systems from RH = 0.9 to 0.3 and then to 0.1.

\begin{table}[tb]
\centering
\small
\caption{Recorded activity during formal production.}
\label{tab:operation_census}
\begin{tabular}{p{3.6cm}p{8.8cm}r}
\toprule
Metric & Definition & Count \\
\midrule
Simulation cycles
& Completed segmented GCMC--MD production runs
& 120 \\

Analysis--decision cycles
& Post-segment analyses followed by a continuation, archive, or review decision
& 120 \\

Accepted states
& System--RH states written to the final authoritative archive
& 15 \\

RH transitions
& Successor RH states initialized from accepted predecessor states
& 10 \\

States requiring review
& Production states for which the approved rules could not safely select the next action
& 1 \\

Reasoning-agent invocations during production
& Non-interactive Codex CLI calls during formal production
& 0 \\
\bottomrule
\end{tabular}
\end{table}

One of the 15 production states required review, but this did not automatically trigger a reasoning-agent call. Formal production contained no non-interactive Codex CLI invocation. The rule-based campaign agent and the scientific computation and analysis tools therefore handled the routine simulation, analysis, archiving, provenance checks, and RH progression throughout the campaign.

This result shows that continuous LLM involvement was not required for this long-running campaign. The reasoning agent was placed at stages where flexible interpretation could be useful, namely campaign construction and review of unresolved states, while routine production remained rule based. For the present campaign, this separation was sufficient to complete 120 simulation cycles and all 15 prescribed states without invoking the reasoning agent during production.

The zero invocation count applies specifically to formal production. It does not include the earlier planning interactions, software development, validation tests, or the two retrospective Codex CLI replays described in Section~3.3.

\subsection{Review Cases and Retrospective Reasoning-Agent Evaluation}

The single production state requiring review was K--LC0.4 at
\(\mathrm{RH}=0.3\). Its accepted \(\mathrm{RH}=0.9\) predecessor ended at an
absolute LAMMPS timestep of \(42.1\times10^{6}\). After restart, the
\(\mathrm{RH}=0.3\) state was stopped at an absolute timestep of
\(60.1\times10^{6}\) because the production limit was incorrectly evaluated
using the inherited absolute timestep. The restart chain and timestep counter
were valid, but the current state had accumulated only
\(18.0\times10^{6}\) RH-local steps. The problem was therefore in the
interpretation of the timestep used by the workflow, rather than in the
simulation trajectory or restart file.

The workflow was corrected to evaluate the production limit using the number
of steps accumulated within the current RH state. K--LC0.4 at
\(\mathrm{RH}=0.3\) was then resumed from the valid
\(60.1\times10^{6}\)-step restart and continued for another
\(6.0\times10^{6}\) steps. The state was accepted at
\(24.0\times10^{6}\) RH-local steps and an absolute timestep of
\(66.1\times10^{6}\). Its accepted restart was subsequently used to initialize
the \(\mathrm{RH}=0.1\) state, which was accepted after a further
\(2.0\times10^{6}\) RH-local steps. Recovery therefore preserved the existing
simulation progress and did not require the preceding state or unrelated
systems to be rerun.

A second case was retained from an earlier development stage to examine
restart provenance. An archive for \(\mathrm{RH}=0.7\) existed and contained
apparently complete output and plausible metadata, but it had been generated
from an earlier smoke-test chain. It was not part of the validated predecessor
chain used for the final campaign, whose prescribed RH path was
\(0.9\rightarrow0.3\rightarrow0.1\). The artifact was therefore rejected as a
source for downstream initialization. This case shows that file existence,
recent modification time, and apparently complete output are not sufficient
to establish that a restart belongs to the accepted campaign history.

The K--LC0.4 timestep case and the historical \(\mathrm{RH}=0.7\) provenance
case were subsequently converted into two frozen retrospective replay cases.
For each replay, the recorded workflow state and supporting evidence were
provided to the Codex CLI reasoning agent, powered by GPT-5.5, while the
original interpretation and subsequent recovery outcome were withheld. Each replay was performed using a frozen case package containing the event description and the evidence available for that incident. The reasoning agent was not provided with the recorded interpretation, subsequent recovery outcome, or the corrected workflow implementation. One
reasoning-agent call was made for each case. The resulting decisions passed
the required output-schema checks. For the timestep case, the reasoning agent
identified the incorrect use of an inherited absolute timestep for an
RH-local production limit and recommended correcting the state-local
accounting before continuation. For the provenance case, it identified the
conflict between the historical artifact and the accepted restart chain and
recommended using the authoritative archive rather than the stale smoke-test
output.

Both replay decisions agreed with the interpretations and actions recorded
for the two cases. These results show that the reasoning agent could interpret
the preserved evidence for these specific examples. Because the evaluation
contained only two cases, however, it does not establish general diagnostic
accuracy across other simulation failures or workflow states.

\subsection{Accepted-State Hydration and Structural Responses}

\begin{figure}[t]
  \centering
  \includegraphics[width=0.55\linewidth]{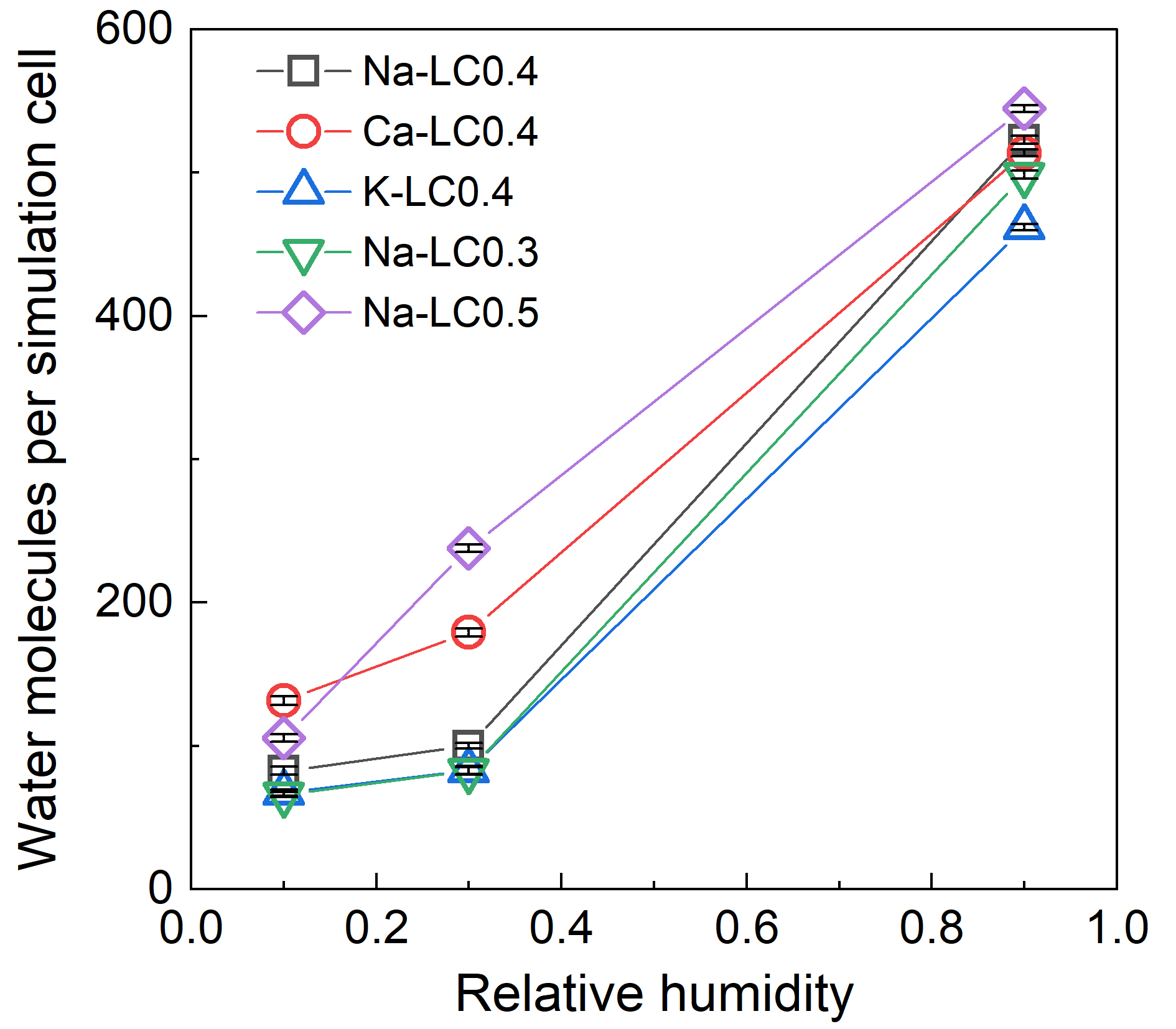}
  \caption{Accepted-state total water populations along the prescribed
  desorption path. Points and error bars denote means and temporal sample
  standard deviations over the final \(1.0\times10^{6}\) MD steps of each
  accepted state, respectively. Water populations are reported per simulation
  cell.}
  \label{fig:total_water_content}
\end{figure}

\begin{figure}[t]
  \centering
  \includegraphics[width=0.95\linewidth]{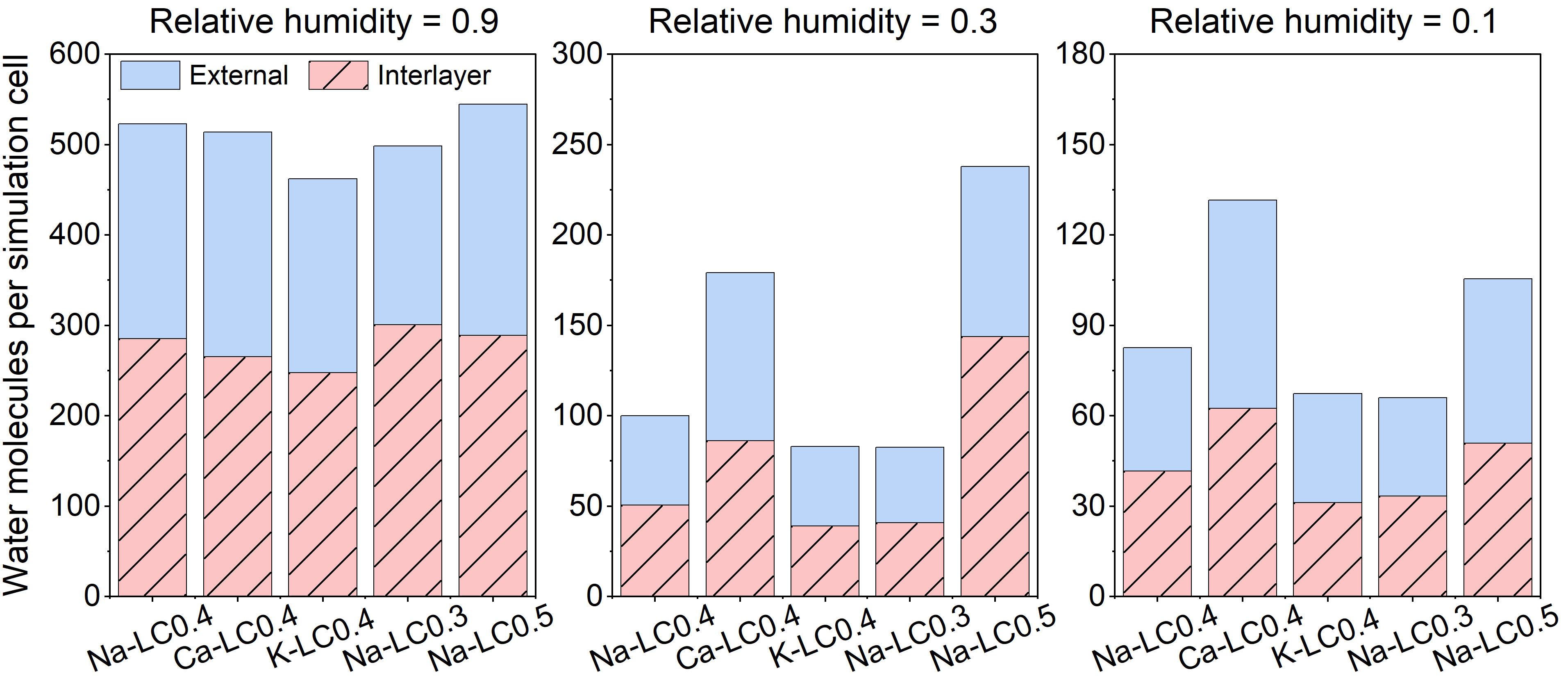}
  \caption{Partitioning of accepted-state water populations between the
  interlayer and external-surface regions. Values are means over the final
  \(1.0\times10^{6}\) MD steps of each accepted state and are reported as
  numbers of water molecules per simulation cell.}
  \label{fig:water_partitioning}
\end{figure}

\begin{figure}[t]
  \centering
  \includegraphics[width=0.55\linewidth]{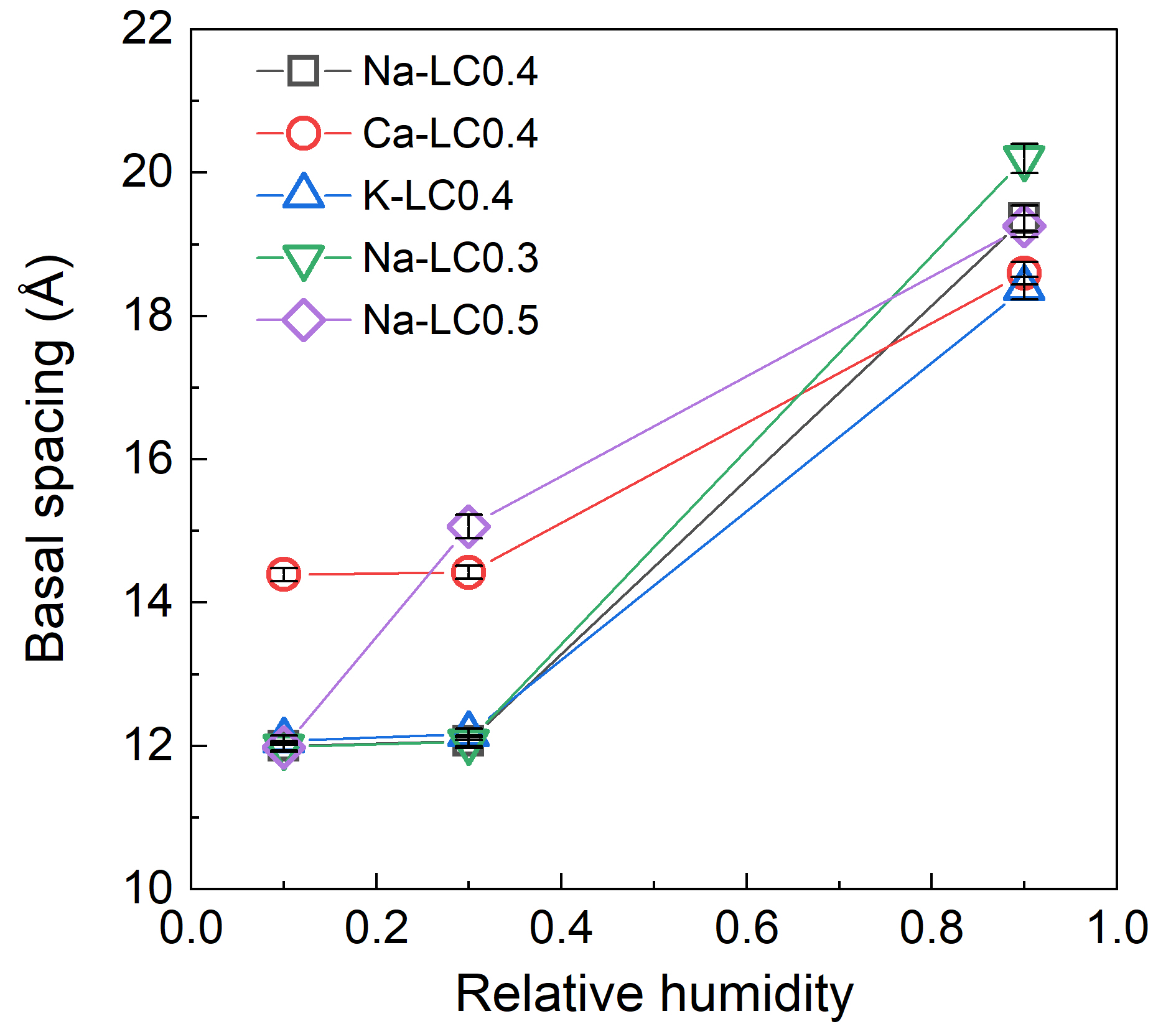}
  \caption{Accepted-state basal spacing along the prescribed desorption path.
  Points and error bars denote means and temporal sample standard deviations
  over the final \(1.0\times10^{6}\) MD steps of each accepted state,
  respectively.}
  \label{fig:basal_spacing_evolution}
\end{figure}

Figures~\ref{fig:total_water_content}--\ref{fig:basal_spacing_evolution}
compare the accepted states obtained after the different RH-local sampling
lengths described in Section~3.1. For each state, the reported value is the
mean over the final \(1.0\times10^{6}\) MD steps. Error bars denote temporal
sample standard deviations over this window and therefore describe
fluctuations within the sampled trajectory rather than uncertainty across
independent replicas. Water populations are reported as numbers of molecules
per simulation cell.

Figure~\ref{fig:total_water_content} shows that all five systems lost
substantial water along the prescribed desorption path. At
\(\mathrm{RH}=0.9\), the systems remained highly hydrated and the differences
among compositions were comparatively limited. The separation became clearer
at \(\mathrm{RH}=0.3\) and \(\mathrm{RH}=0.1\), indicating that cation
composition and layer-charge-related differences became more important as
water availability decreased.

Among the three LC0.4 systems, Ca--LC0.4 retained the greatest total water
population at both lower RH values. Na--LC0.4 showed intermediate retention,
whereas K--LC0.4 reached the driest accepted state at
\(\mathrm{RH}=0.1\). This ordering is consistent with differences in cation
hydration and interlayer association reported previously. In particular,
Ca\(^{2+}\) can maintain a comparatively stable hydration shell under dry
conditions, whereas K\(^{+}\) more readily forms compact inner-sphere
interlayer configurations \cite{zhang2014hydration,wang2025ca,yotsuji2021cation}.
The present total-water results are consistent with these mechanisms but do
not alone resolve the coordination environment of the cations.

Within the Na series, Na--LC0.5 retained substantially more water than
Na--LC0.3 and Na--LC0.4 at \(\mathrm{RH}=0.3\) and
\(\mathrm{RH}=0.1\). However, the number of charge-balancing Na\(^{+}\) ions
also increased from 12 in Na--LC0.3 to 16 in Na--LC0.4 and 20 in
Na--LC0.5. The observed trend therefore represents the combined variation of
nominal layer charge and compensating-ion population rather than an isolated
layer-charge effect at fixed ion number. Previous molecular simulations have
shown that layer charge affects water adsorption, swelling transitions, and
the distribution and mobility of interlayer and external-surface
species \cite{tambach2004swelling,yang2019layercharge}. The present results
similarly show a composition-dependent resistance to dehydration, but they do
not separate the individual contributions of layer charge and Na\(^{+}\)
population.

\FloatBarrier

Figure~\ref{fig:water_partitioning} separates the total water population into
interlayer and external-surface contributions. Both populations generally
decreased with RH, but their relative changes differed among systems. The
composition-dependent differences in Figure~\ref{fig:total_water_content}
therefore arose not only from different total water populations but also from
different distributions of the remaining water.

For the LC0.4 cation comparison, the greater low-RH water retention of
Ca--LC0.4 was associated with both interlayer and external-surface water, with
a particularly clear difference in the interlayer population. At
\(\mathrm{RH}=0.3\) and \(\mathrm{RH}=0.1\), Ca--LC0.4 retained more
interlayer water than Na--LC0.4 and K--LC0.4, whereas K--LC0.4 retained the
smallest interlayer population. The total-water ordering in
Figure~\ref{fig:total_water_content} therefore reflects a substantial
cation-dependent difference in residual interlayer hydration. This behavior
is consistent with the stronger hydration of Ca\(^{2+}\) and the reported
dependence of montmorillonite swelling on cation hydration free energy and
inner- or outer-sphere association \cite{zhang2014hydration,wang2025ca,yotsuji2021cation}.

The Na series showed a related but not identical response. At
\(\mathrm{RH}=0.3\), Na--LC0.5 retained more water in both the interlayer and
external-surface regions than Na--LC0.3 and Na--LC0.4, with the largest
difference occurring in the interlayer. Na--LC0.5 also remained the most
interlayer-hydrated Na system at \(\mathrm{RH}=0.1\), although the difference
was smaller than at \(\mathrm{RH}=0.3\). These results indicate that the
greater total retention of Na--LC0.5 was not caused solely by additional
external adsorption. It also involved stabilization of a larger residual
interlayer-water population. As noted above, this response must be attributed
to the combined change in nominal layer charge and charge-balancing Na\(^{+}\)
population.

\FloatBarrier

The basal-spacing response in Figure~\ref{fig:basal_spacing_evolution}
provides the corresponding structural comparison. Desorption generally
reduced both the interlayer-water population and the basal spacing, showing
that loss of interlayer water was accompanied by clay-sheet contraction.
However, the relation was not one-to-one across all systems and RH conditions.

For the LC0.4 cation series, all three systems were expanded at
\(\mathrm{RH}=0.9\), but their structures differed clearly after drying.
Na--LC0.4 and K--LC0.4 reached compact basal spacings of approximately
\(12~\text{\AA}\) at both \(\mathrm{RH}=0.3\) and
\(\mathrm{RH}=0.1\). In contrast, Ca--LC0.4 remained expanded at approximately
\(14.4~\text{\AA}\). Together with
Figure~\ref{fig:water_partitioning}, this result shows that the larger
low-RH basal spacing of Ca--LC0.4 coincided with its larger retained
interlayer-water population. The response is consistent with stable
Ca\(^{2+}\) hydration supporting an expanded interlayer, whereas the compact
K-bearing state is consistent with the preference of K\(^{+}\) for
inner-sphere configurations \cite{wang2025ca,yotsuji2021cation}. Basal spacing
alone, however, is insufficient to assign a unique ion coordination or
hydration-layer structure.

Within the Na series, Na--LC0.5 maintained a basal spacing of approximately
\(15~\text{\AA}\) at \(\mathrm{RH}=0.3\), while Na--LC0.3 and Na--LC0.4
contracted to approximately \(12~\text{\AA}\). This structural difference
coincided with the substantially larger interlayer-water population of
Na--LC0.5 and is consistent with layer-charge-dependent changes in adsorption
and swelling behavior reported in previous molecular simulations \cite{tambach2004swelling,teichmcgoldrick2015swelling,yang2019layercharge}.

The relationship nevertheless changed at other RH conditions. At
\(\mathrm{RH}=0.9\), Na--LC0.3 had the largest basal spacing among the Na
systems even though Na--LC0.5 contained more total water. At
\(\mathrm{RH}=0.1\), all three Na systems reached similar compact basal
spacings despite retaining different water populations. These comparisons
show that basal spacing is controlled more directly by the amount and
organization of interlayer water than by the total water population alone.
External-surface water contributes to the total uptake without necessarily
increasing clay-sheet separation, while ion position and molecular
organization can permit different water populations within similar basal
spacings.

\FloatBarrier

Taken together, Figures~\ref{fig:total_water_content}--%
\ref{fig:basal_spacing_evolution} show three related but non-equivalent
responses. Total water content describes how much water remained, water
partitioning identifies whether it remained in the interlayer or on the
external surfaces, and basal spacing records the associated structural
response of the clay sheets. Within the investigated systems, exchangeable
cation identity and the coupled variation of layer charge and
charge-balancing Na\(^{+}\) population affected all three responses, with the
largest compositional differences emerging after drying to
\(\mathrm{RH}=0.3\) and \(\mathrm{RH}=0.1\).
\section{Discussion}

The completed campaign shows that a long-running simulation workflow does not require continuous LLM involvement. The reasoning agent was used during campaign construction and remained available for states that could not be handled safely by the approved rules, while routine production was managed by the rule-based campaign agent. This arrangement was sufficient for the present campaign to complete the prescribed simulations, analyses, archives, and RH transitions without invoking the reasoning agent during formal production. The absence of production-stage LLM calls should therefore not be interpreted as an absence of agent support. Rather, it indicates that reasoning was placed where interpretation could be useful instead of being repeated during every routine operation.

Reasoning agents can improve the efficiency of scientific work by helping researchers translate scientific objectives into executable plans and by examining unusual states that would otherwise require manual investigation. However, using a reasoning agent for every simulation segment would also repeat many decisions that are already defined by the campaign rules. It would require repeated transmission of workflow context, increase token use and latency, and introduce model variability into actions that can be performed reproducibly by conventional code. For routine continuation, analysis, archiving, and state progression, a rule-based agent is therefore more suitable. The present study did not directly measure the cost or speed difference between selective and continuous LLM use, but it demonstrates that continuous reasoning was unnecessary for this campaign.

This distinction also suggests that scientific software should provide a working path designed specifically for agents. Existing simulation programs are mainly designed for human users, who can inspect directories, read log files, remember earlier decisions, and resolve ambiguous filenames from context. A reasoning agent should not be expected to reconstruct this hidden information repeatedly or to operate a human-oriented interface through open-ended interpretation. Instead, the workflow should expose the current simulation state, valid restart source, completed analyses, permitted next actions, and predecessor--successor relationships in machine-readable form. Scientific tools should accept structured inputs and return structured results, while important actions should remain visible in human-readable records. Designing this separate agent-oriented path is more useful than simply placing an LLM around an existing collection of scripts and files.

The reviewed cases also show why some decisions cannot be reduced to checking whether a simulation has crashed or reached a numerical threshold. A recorded value can be numerically correct but used for the wrong purpose, and an apparently complete archive can still be unsuitable because it does not belong to the accepted restart chain. When the approved rules cannot resolve such a conflict, the workflow should stop before the questionable state is propagated downstream. A reasoning agent can then examine a structured event and its supporting evidence and recommend an action. That recommendation should not be executed directly: it must first pass the specified schema and policy checks and, when scientific judgment is required, human review. The two retrospective replays show that the reasoning agent produced appropriate responses for the two preserved cases, but they do not establish general failure-diagnosis capability.

Several limitations define the scope of these conclusions. The study considered one GCMC--MD campaign, one formal-production state requiring review, and two retrospective replay cases. It did not compare selective reasoning with continuous LLM control or human-only review in terms of computational cost, response time, or recovery quality. In addition, the scientific objective, molecular models, RH path, monitored observables, acceptance rules, review triggers, and permitted responses were defined or approved by the researcher. The reasoning agent did not independently establish the physical validity of the force field or simulation protocol. The reliability of Agent-MD therefore still depends on the quality of the scientific model, the approved rules, the recorded state, and the evidence supplied for review.

The same approach could be applied to other computational studies when execution can be divided into restartable stages, routine decisions can be written as explicit rules, and the origin of each successor state can be recorded. Each application would require its own definitions of state, acceptance, provenance, and conditions requiring review. Further work should test the approach in additional production campaigns with a broader range of genuine incidents and compare selective reasoning, continuous LLM involvement, and human review using measurable criteria such as cost, latency, recovery correctness, and reproducibility.
\section{Conclusions}

This work introduced Agent-MD, a workflow for combining LLM-based reasoning with rule-based execution in long-running molecular-simulation campaigns. The reasoning agent was used to help construct the campaign and to review states that could not be resolved safely by the approved rules. Routine simulation, analysis, continuation, archiving, provenance checking, and progression between prescribed conditions were managed by a persistent rule-based campaign agent. This separation avoided repeated LLM use for actions that could be performed more directly and reproducibly by conventional code.

Agent-MD was applied to a GCMC--MD study of water-vapor desorption in five montmorillonite systems along the prescribed path
\(\mathrm{RH}=0.9\rightarrow0.3\rightarrow0.1\). All 15 system--RH states were accepted and archived after 120 segmented simulation cycles. Different states required substantially different sampling lengths, showing the value of reassessing each state rather than assigning a uniform production length. Formal production required no non-interactive reasoning-agent call, although one state was paused for review when the workflow incorrectly applied an inherited absolute timestep to an RH-local sampling limit. Two preserved cases were later replayed through the reasoning agent, which produced decisions consistent with the recorded interpretations for those specific cases.

The completed campaign also produced clear composition-dependent hydration and structural responses. At low RH, Ca--LC0.4 retained more interlayer water and maintained a larger basal spacing, whereas K--LC0.4 underwent stronger dehydration and contraction. Within the Na series, Na--LC0.5 retained more water under dry conditions, although this comparison reflects the combined variation of layer charge and the corresponding charge-balancing Na\(^{+}\) population. The results further showed that total water content, water partitioning, and basal spacing provide related but non-equivalent descriptions of the desorption response.

The present study demonstrates the feasibility of selective reasoning-agent involvement in one stateful GCMC--MD campaign; it does not establish general diagnostic performance across molecular-simulation problems. Broader evaluation will require additional production campaigns, more genuine review cases, and direct comparisons with continuous LLM use and human-only review. More generally, effective scientific agents will require workflows designed for their use, with explicit state, restart provenance, permitted actions, and structured evidence, rather than repeated interpretation of human-oriented scripts and file systems.
\section*{Acknowledgments}

This work was supported by the Research Grants Council (RGC) of the Hong Kong Special Administrative Region Government (HKSARG) of China under Grant Nos. 15231825, N\_PolyU534/20, 15217220.

\section*{Code and Data Availability}

The Agent-MD source code, campaign specifications, analysis and plotting
scripts, processed data underlying the reported tables and figures, and the
two frozen replay cases are available at
\url{https://github.com/yj-wang14/Agent-MD}. The repository also contains the
accepted-state manifest and provenance records used to identify the final
scientific results and to exclude test, stale, failed, or superseded outputs.

\bibliography{references}

\end{document}